\documentclass[10pt, conference, compsocconf]{IEEEtran}
\ifCLASSINFOpdf
\else
\fi

\usepackage{caption}

\usepackage{nopageno}
\usepackage{multirow}
\usepackage{graphics}
\usepackage{algpseudocode}
\usepackage{pseudocode}
\usepackage{algorithm}
\usepackage{latexsym}
\usepackage{graphicx}
\usepackage{subcaption}
\usepackage{mathptmx}
\usepackage{amsmath}
\usepackage{array}
\usepackage{url}
\usepackage{color}
\usepackage{dblfloatfix}

\usepackage{hyperref}

\makeatletter
\def\footnoterule{\kern-3\p@
  \hrule \@width 2in \kern 2.6\p@} 
\makeatother

\newcommand{\squeezeup}{\vspace{-3mm}}
\renewcommand{\baselinestretch}{0.935}

\newcommand{\sys}{ReBNet}

\newcolumntype{L}[1]{>{\raggedright\let\newline\\\arraybackslash\hspace{0pt}}m{#1}}
\newcolumntype{C}[1]{>{\centering\let\newline\\\arraybackslash\hspace{0pt}}m{#1}}
\newcolumntype{R}[1]{>{\raggedleft\let\newline\\\arraybackslash\hspace{0pt}}m{#1}}

\begin{document}

%


%

%
\title{ReBNet: Residual Binarized Neural Network}


\author{\IEEEauthorblockN{Mohammad Ghasemzadeh, Mohammad Samragh, and Farinaz Koushanfar}
\IEEEauthorblockA{Department of Electrical and Computer Engineering, University of California San Diego\\
\{mghasemzadeh, msamragh, farinaz\}@ucsd.edu}}


\maketitle

\begin{abstract}

This paper proposes ReBNet, an end-to-end framework for training reconfigurable binary neural networks on software and developing efficient accelerators for execution on FPGA. Binary neural networks offer an intriguing opportunity for deploying large-scale deep learning models on resource-constrained devices. Binarization reduces the memory footprint and replaces the power-hungry matrix-multiplication with light-weight XnorPopcount operations. However, binary networks suffer from a degraded accuracy compared to their fixed-point counterparts. We show that the state-of-the-art methods for optimizing binary networks accuracy, significantly increase the implementation cost and complexity. To compensate for the degraded accuracy while adhering to the simplicity of binary networks, we devise the first reconfigurable scheme that can adjust the classification accuracy based on the application. Our proposition improves the classification accuracy by representing features with \emph{multiple} levels of residual binarization. Unlike previous methods, our approach does not exacerbate the area cost of the hardware accelerator. Instead, it provides a tradeoff between throughput and accuracy while the area overhead of multi-level binarization is negligible.


\thispagestyle{empty}

\end{abstract}

%
\IEEEpeerreviewmaketitle

\section{Introduction}

Convolutional Neural Networks (CNNs) are widely used in a variety of machine learning applications, many of which are deployed on embedded devices~\cite{li2017deeprebirth,zhang2017shufflenet,howard2017mobilenets}. With the swarm of emerging intelligent applications, development of real-time and low-power hardware accelerators is especially critical for resource-limited settings. A line of research has therefore been focused on the development of FPGA accelerators for execution of CNN applications~\cite{zhang2015optimizing,ovtcharov2015accelerating,suda2016throughput}. Although the building blocks of CNNs are highly parallelizable, the high computational complexity and memory footprint of these models are barriers to efficient implementation. 

A number of prior works have focused on reducing the computational complexity and memory footprint of CNNs by trimming the redundancies of the model prior to designing an accelerator. Examples of such optimization techniques include tensor decomposition~\cite{kim2015compression,zhang2015efficient,nazemi2018hardware}, parameter quantization~\cite{hubara2016quantized,han2015deep,samragh2017customizing}, sparse convolutions~\cite{liu2015sparse,wen2016learning}, and training binary neural networks~\cite{courbariaux2016binarized,rastegari2016xnor}.

Among the above optimization techniques, binary networks result in two particular benefits: (i) They reduce the memory footprint compared to models with fixed-point parameters; this is especially important since memory access plays an essential role in the execution of CNNs on resource-constrained platforms. (ii) Binary CNNs replace the power-hungry multiplications with simple XNOR operations~\cite{rastegari2016xnor,umuroglu2017finn}, significantly reducing the runtime and energy consumption. Consequent to the aforementioned benefits, the  dataflow architecture of binary CNNs is remarkably simpler than their fixed-point counterparts.

There are several remaining challenges for training binary CNNs. First, the training phase of binary neural networks is often slow and the final classification accuracy is typically lower than the model with full-precision parameters. It has been shown that the loss of accuracy can be partially evaded by training binary networks that have wider layers~\cite{umuroglu2017finn}. Nevertheless, this method for accuracy enhancement diminishes the performance gains of binarization. In another effort, authors of XNOR-net~\cite{rastegari2016xnor} tried to improve the accuracy of the binary CNN using scaling factors that are computed by averaging the features during inference. This method, however, sacrifices the simplicity of the binary CNN accelerator by adding extra full-precision calculations to the computation flow of binary CNN layers. Similarly, the approach in~\cite{tang2017train} involves multiple rounds of computing the average absolute value of input activations which incurs an excessive computation cost during the inference phase. Ternary neural networks~\cite{alemdar2017ternary,mellempudi2017ternary} may surpass binary CNNs in terms of the inference accuracy; however, they are deprived of the benefits of simple XNOR operations.

We argue that a practical solution for binary CNNs should possess two main properties: (i)~The accuracy of the binary model should be comparable to its full-precision counterpart. (ii)~The proposed method to improve the accuracy should not compromise the low overhead and accelerator scalability benefits of binary networks. This paper proposes \sys{}, an end-to-end framework for training reconfigurable binary CNNs in software and developing efficient hardware accelerators for execution on FPGA. We introduce the novel concept of multi-level binary CNNs and design algorithms for learning such models. Building upon this idea, we design and implement a scalable FPGA accelerator for binary CNNs. The benefit of our approach over existing binarization methods is that the number of binarization levels can be adjusted for different applications without significant hardware modification.

In \sys{}, the weight parameters of CNN layers are represented with 1-level binarized values, while a multi-level residual binarization scheme is learned for the activation units. As such, the memory footprint of the \emph{parameters} in \sys{} is the same as that of a single-level Binary CNN. We show that the accuracy of \sys{} can be improved by using 2 or 3 levels of residual binarization with a negligible area overhead. The design of \sys{} residual binarization is compatible with the standard $XnorPopcount$ operations; the underlying computations involving the feature vectors can be decomposed into a number of standard $XnorPopcount$ operations. As such, \sys{} provides scalability for the design of binary CNNs. The contributions of this paper are summarized as follows:
\begin{itemize}
    \item Proposing residual binarization as a reconfigurable dimension of binary CNNs. We devise an activation function with few scaling factors that are used for residual binarization. We also introduce a method for training the scaling factors. 
    \item Development of an Application Programming Interface (API) for training multi-level binarized CNNs\footnote{Codes are available at~\url{https://github.com/mohaghasemzadeh/ReBNet}\label{xx}}.
    \item Creation of a hardware library for implementation of different CNN layers using \sys{} methodology. The library allows users to configure the parallelism in each CNN layer using high-level parameters\textsuperscript{\ref{xx}}.
    \item Performing proof-of-concept evaluations on four benchmarks on three FPGA platforms.
\end{itemize}

\section{Preliminaries}\label{sec:prelim}
In this section, we outline the operations of binary CNNs and their hardware implementation. Specifically, we briefly describe the FPGA design proposed by~\cite{umuroglu2017finn}. Please refer to the original paper for a more detailed explanation.

\subsection{Binary CNN Operations}

Neural networks are composed of multiple convolution, fully-connected, activation, batch-normalization, and max-pooling layers. Binarization enables the use of a simpler equivalent for each layer as explained in this section. 

\begin{figure}
    \centering
    \includegraphics[width=0.7\columnwidth]{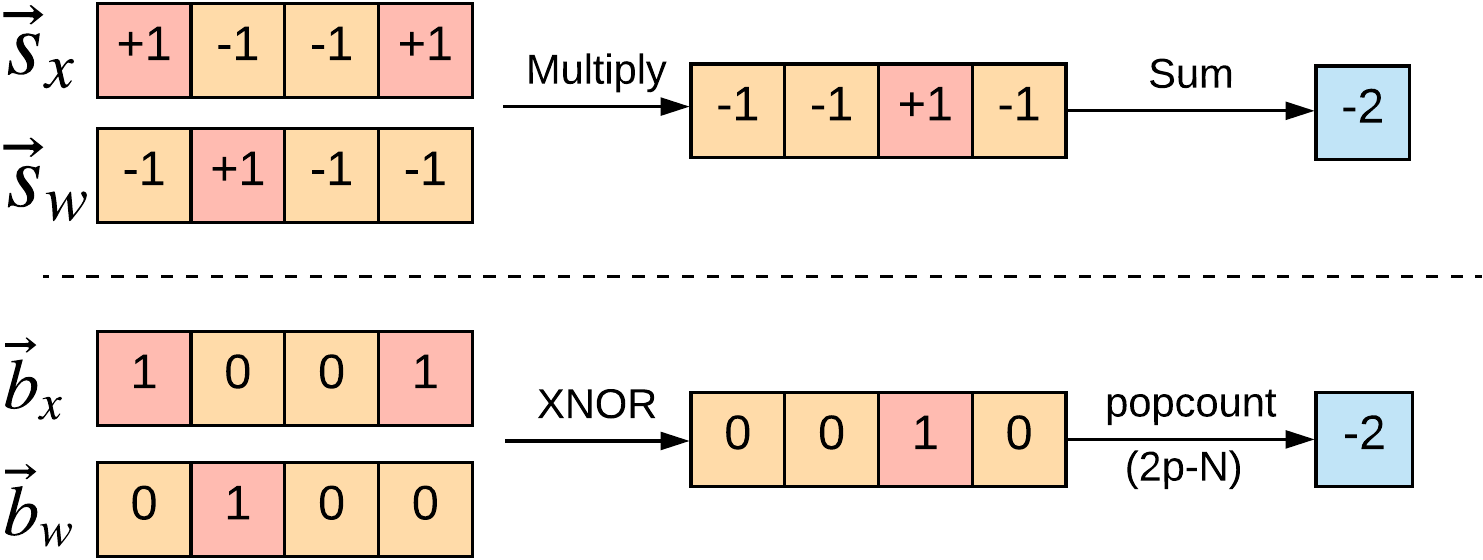}
    \caption{The equivalence of dot product (top) and $XnorPopcount$ (bottom) operations. In the popcount operation, $p$ is the number of set bits and $N$ is the size of the vector.}
    \label{fig:xnorpopcount}
\end{figure}

\noindent{\bf Binary dot product:}
The computational complexity of neural networks is mostly owing to the convolution and fully-connected layers. Both layers can be broken into a number of dot products between input features and weight parameters. A dot product accumulates the element-wise products of a feature vector $\vec{x}$ and a weight vector $\vec{w}$:
\begin{equation}
    dot(\vec{x}, \vec{w}) = \sum_i \vec{x}[i]\times \vec{w}[i]
\end{equation}
In case of binary CNNs, the elements of $\vec{x}$ and $\vec{w}$ are restricted to binary values $\pm \gamma_x$ and $\pm \gamma_w$, respectively. The dot product of these vectors can be efficiently computed using $XnorPopcount$ operations as suggested in~\cite{courbariaux2016binarized,rastegari2016xnor}. Let $\vec{x}=\gamma_x\ \vec{s}_x$ and $\vec{w}=\gamma_w\ \vec{s}_w$, where $\{\gamma_x,\gamma_w\}$ are scalar values and $\{\vec{s}_x,\vec{s}_w\}$ are sign vectors whose elements are either $+1$ or $-1$. If we encode the sign values ($-1\rightarrow 0$ and $+1\rightarrow 1$), we obtain binary vectors $\{\vec{b}_x,\vec{b}_w\}$. The dot product between $\vec{x}$ and $\vec{w}$ can be computed as:
\begin{equation}
\resizebox{0.9\columnwidth}{!}{
    $dot(\vec{w},\vec{x})\ = \gamma_x \gamma_w\ dot(\vec{s}_x,\vec{s}_w)=\gamma_x \gamma_w\ XnorPopcount(\vec{b_{x}},\vec{b_{w}}$)
    }
\end{equation}
Figure~\ref{fig:xnorpopcount} depicts the equivalence of $dot(\vec{s}_x,\vec{s}_w)$ and $XnorPopcount(\vec{b_{x}},\vec{b_{w}})$ using an example. 

\noindent{\bf Binary activation:}
The binary activation function encodes an input $y$ using a single bit based on the sign of $y$. Therefore, the hardware implementation only requires a comparator.

\noindent{\bf Binary batch-normalization:}
It is often useful to normalize the result of the dot product $y=dot(\vec{x},\vec{w})$ before feeding it to the binary activation function described above. A batch-normalization layer converts each input $y$ into $\alpha \times y - \beta$, where $\alpha$ and $\beta$ are the parameters of the layer. Authors of~\cite{umuroglu2017finn} suggest combining batch-normalization and binary-activation layers into a single thresholding layer. The cascade of the two layers computes the following:
\begin{equation}
   output= Sign(\alpha\times y - \beta)= Sign(y - \frac{\beta}{\alpha}) 
\end{equation}
Therefore, the combination of the two layers only requires a comparison with the threshold value $\frac{\beta}{\alpha}$.

\begin{figure}
    \centering
    \includegraphics[width=0.9\columnwidth]{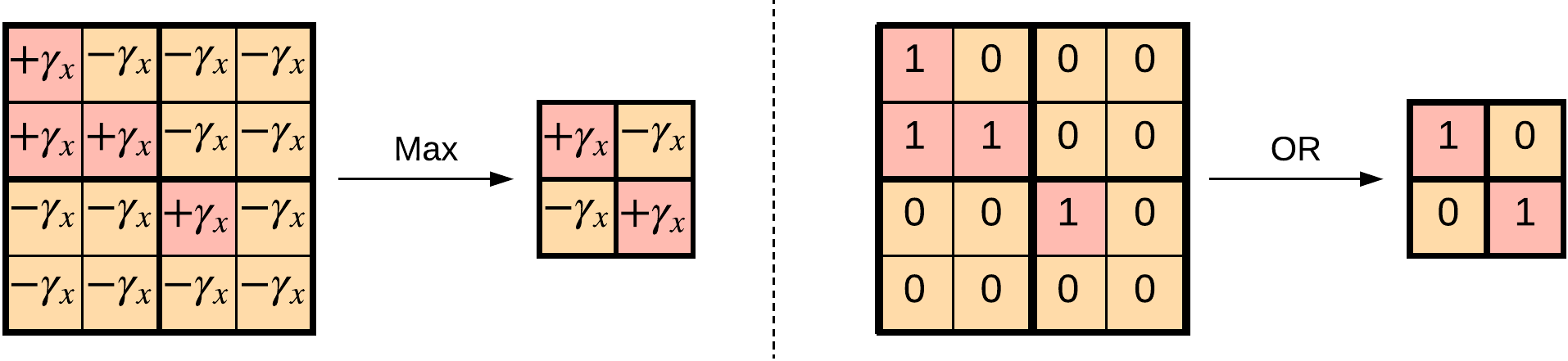}
    \caption{The equivalence of max-pooling operation over binarized features (left) and OR operation over encoded features (right).}
    \label{fig:max-pool}
    \squeezeup
    \squeezeup
\end{figure}

\noindent{\bf Binary max-pooling:}
A max-pooling layer computes the maximum of features over sliding windows of the input feature map. The max-pooling operation can be performed by element-wise OR over the binary features. Figure~\ref{fig:max-pool} depicts the equivalence between max-pooling over fixed-point features $\vec{x}\in\{+\gamma,-\gamma\}^N$ and element-wise OR over binary features $\vec{b}_x\in\{0,1\}^N$.

\subsection{Hardware Implementation and Parallelism}\label{sec:prelim-hw}

\begin{figure}
    \centering
        \begin{subfigure}[b]{0.85\columnwidth}
                \includegraphics[width=\columnwidth]{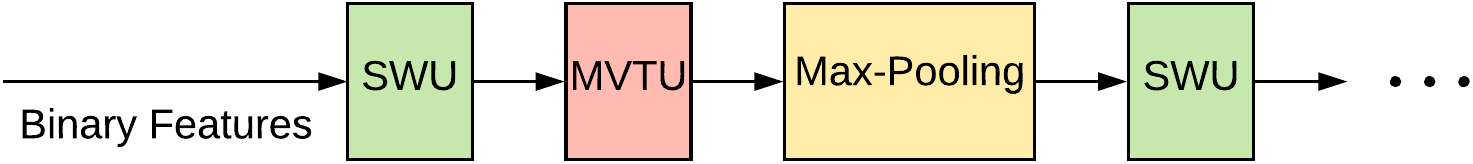}
                \caption{}
        \end{subfigure}
        \vspace{1em}
        \begin{subfigure}[b]{0.85\columnwidth}
                \includegraphics[width=\columnwidth]{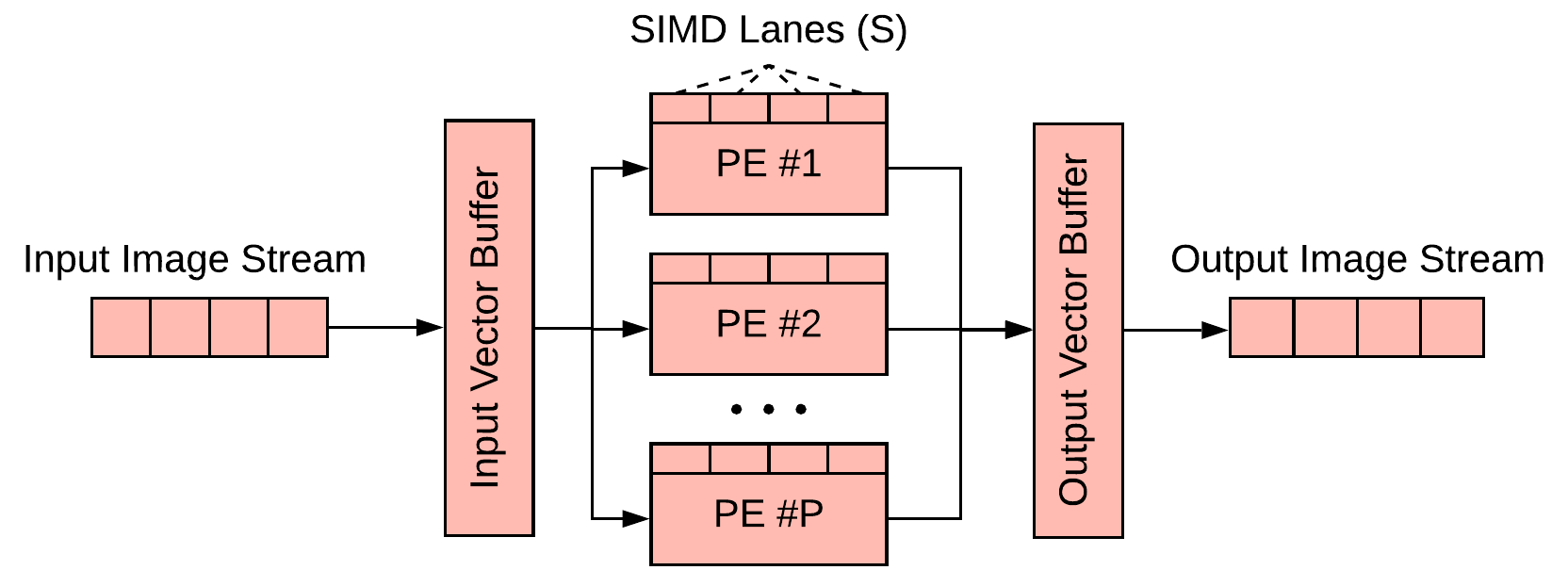}
                \caption{}
        \end{subfigure}
        \vspace{1em}
        \begin{subfigure}[b]{0.85\columnwidth}
                \includegraphics[width=\columnwidth]{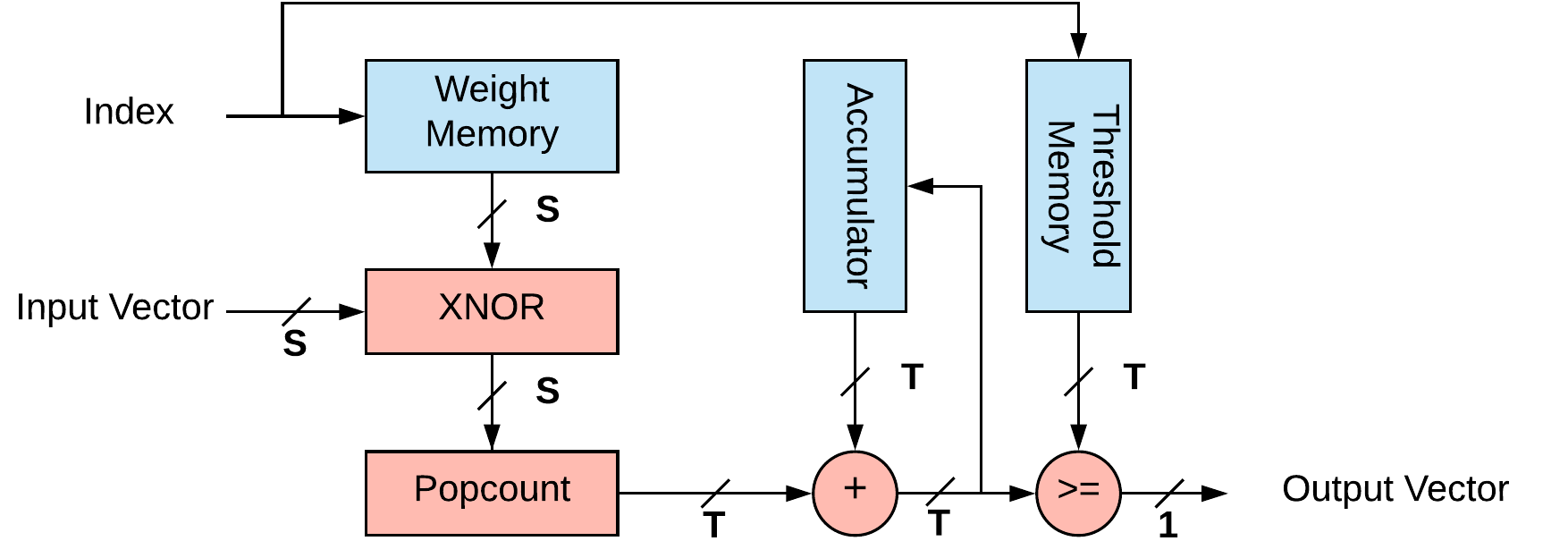}
                \caption{}
        \end{subfigure}
        \caption{(a)~Computation flow of FINN~\cite{umuroglu2017finn} accelerator. (b)~An MVTU with ``P'' processing elements, each with SIMD-width of ``S''. (c)~Architecture of a processing unit. ``S'' is the SIMD-width and ``T'' is the fixed-point bitwidth.}
        \label{fig:base_accelerator}
       
\end{figure}

\noindent{\bf Matrix multiplication:}
Authors of~\cite{umuroglu2017finn} propose the flow diagram of Figure~\ref{fig:base_accelerator}-(a) to implement different layers of binary CNNs on FPGA. The sliding window unit (SWU) scans the input feature maps of convolution layers and feeds appropriate values to the corresponding matrix vector threshold unit (MVTU). Both convolution and fully-connected layers are implemented using the MVTU, which realizes matrix-vector multiplication, batch normalization, and binary activation. 

\noindent{\bf Parallelism:}
The MTVU offers two levels of parallelism for matrix-vector multiplication as depicted in Figure~\ref{fig:base_accelerator}-(b,c). First, each MTVU has a number of processing elements (PEs) that compute multiple output neurons in parallel; each PE is responsible for a single dot product between two binarized vectors. Second, each PE breaks the corresponding operation into a number of sub-operations, each of which performed on SIMD-width binary words. 

\section{Overview}
\begin{figure}
    \centering
    \includegraphics[width=0.75\columnwidth]{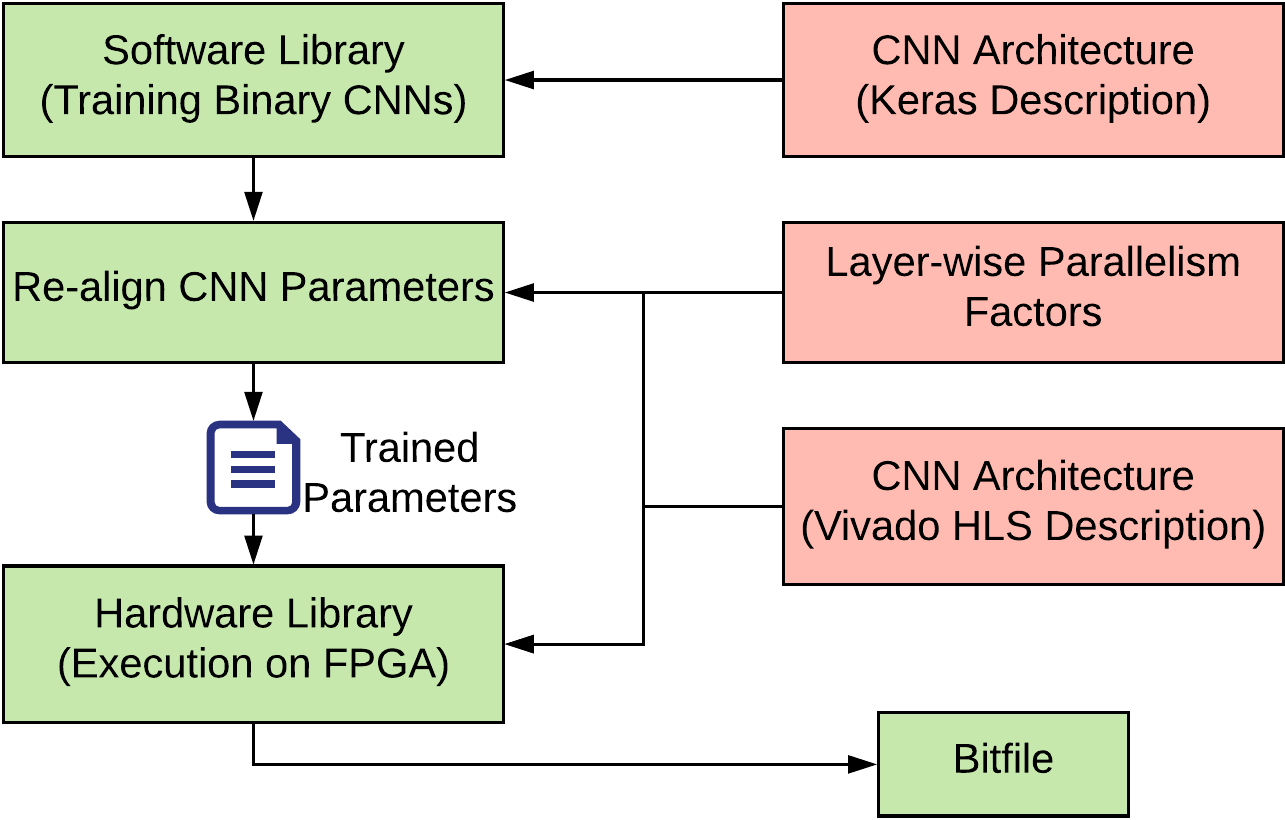}
    \caption{The global flow of \sys{} software training and hardware accelerator synthesis. }
    \label{fig:global}
    \vspace{-0.55em}
\end{figure}

The global flow of \sys{} API is presented in Figure~\ref{fig:global}. The user provides the CNN architecture to the software library, which trains the binary CNN with a specified number of residual binarization levels. She/he also provides the network description using our hardware library, along with the parallelism factors for the hardware accelerator. Based on these parallelism factors (PE-count and SIMD-width), the binary network parameters are re-aligned and stored appropriately to be loaded into the hardware accelerator. The bitfile is then generated using the hardware library.

In this section, we first describe residual binarization as a reconfigurable dimension in the design of binary CNNs. We next discuss the training methodology for residual binary networks. Finally, we elaborate on our hardware accelerator.

\subsection{Residual Binarization}\label{sec:resid}

Imposing binary constraints on weights and activations of a neural network limits the model's ability to provide the inference accuracy that a floating-point or fixed-point counterpart would achieve. To address this issue, we propose a multi-level binarization scheme where the residual errors are sequentially binarized to increase the numerical precision of the approximation.

\begin{figure}
    \centering
    \includegraphics[width=\columnwidth]{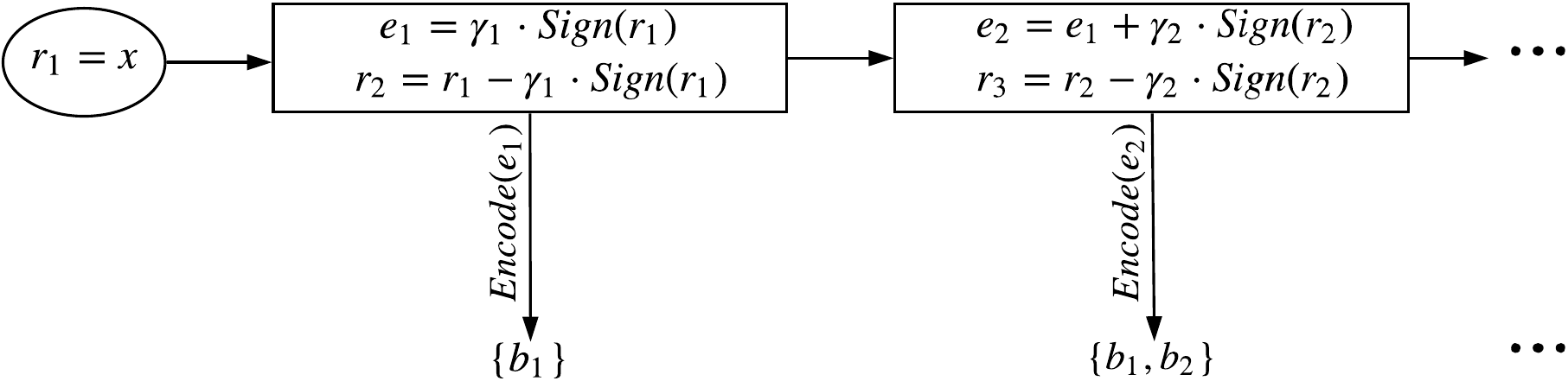}
    \caption{Schematic flow for computing an $M$-level residual binary approximate $e_M$ and the corresponding encoded bits $\{b_1, b_2, \cdots,b_M\}$. As one goes deeper in levels, the estimation becomes more accurate. In this figure, we drop the subscript $x$ from $\gamma_{xi}$ and represent them as $\gamma_i$ for simplicity.}
    \label{fig:resid}
    \squeezeup
    \squeezeup
\end{figure}

\noindent{\bf Multi-level residual binarization}:  Figure~\ref{fig:resid} presents the procedure to approximate a fixed-point input $x$ with a multi-level binarized value $e$. For an $M$-level binarization scheme, there exist $M$ scaling factors $\{\gamma_1,\gamma_2,\dots,\gamma_M\}$. The first level of binarization takes input $x$ and approximates it by either $+\gamma_1$ or $-\gamma_1$ based on the sign of $x$, then computes the residual error $r_2=x-\gamma_1\cdot Sign(x)$. The second level of binarization approximates $r_2$ by either $+\gamma_2$ or $-\gamma_2$ based on the sign of $r_2$, then computes the residual error $r_3=r_2-\gamma_2\cdot Sign(r2)$. Repeating this process for $M$-times results in an $M$-level binarization for the input value. More formally, an input $x$ can be approximated as $e=\sum_{i=1}^{M}\gamma_i\cdot Sign(r_i)$ with $r_i$ being the $i$-th residual error. In \sys{}, the same set of scaling factors is used for all features corresponding to a certain CNN layer; therefore, the features can be encoded using $M$ bits. Algorithm~\ref{alg:resenc} presents the procedure for computing encoded bits $\{b_1, b_2, \cdots, b_n\}$.

\begin{algorithm}
\small
\caption{$M$-level residual encoding algorithm}\label{alg:resenc}
\textbf{inputs:}{$\ \ \gamma_1, \gamma_2, ..., \gamma_M$}, $x$
\\
\textbf{outputs:}{$\ \ b_1, b_2, ..., b_M$}
\vspace{0.2em}
\\\rule {245pt}{0.5pt} 
\begin{algorithmic}[1]
\State $r \gets x$
\For {$i=1 \dots M$}
\State $b_i \gets Binarize(Sign(r))$
\State $r \gets r-Sign(r)\times \gamma_i$
\EndFor
\end{algorithmic}
\end{algorithm}

\begin{figure}
        \centering
        \begin{subfigure}[b]{0.35\columnwidth}
                \includegraphics[width=\columnwidth]{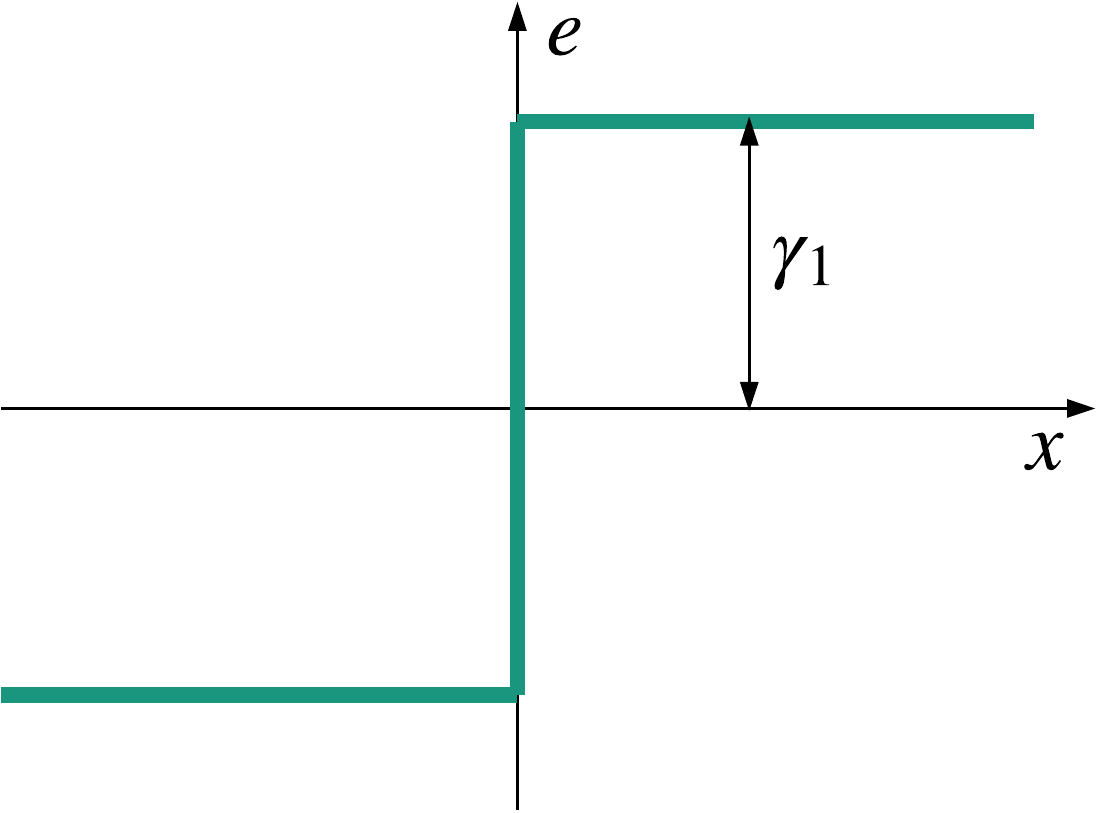}
                \caption{}
        \end{subfigure}
        ~
        \begin{subfigure}[b]{0.35\columnwidth}
                \includegraphics[width=\columnwidth]{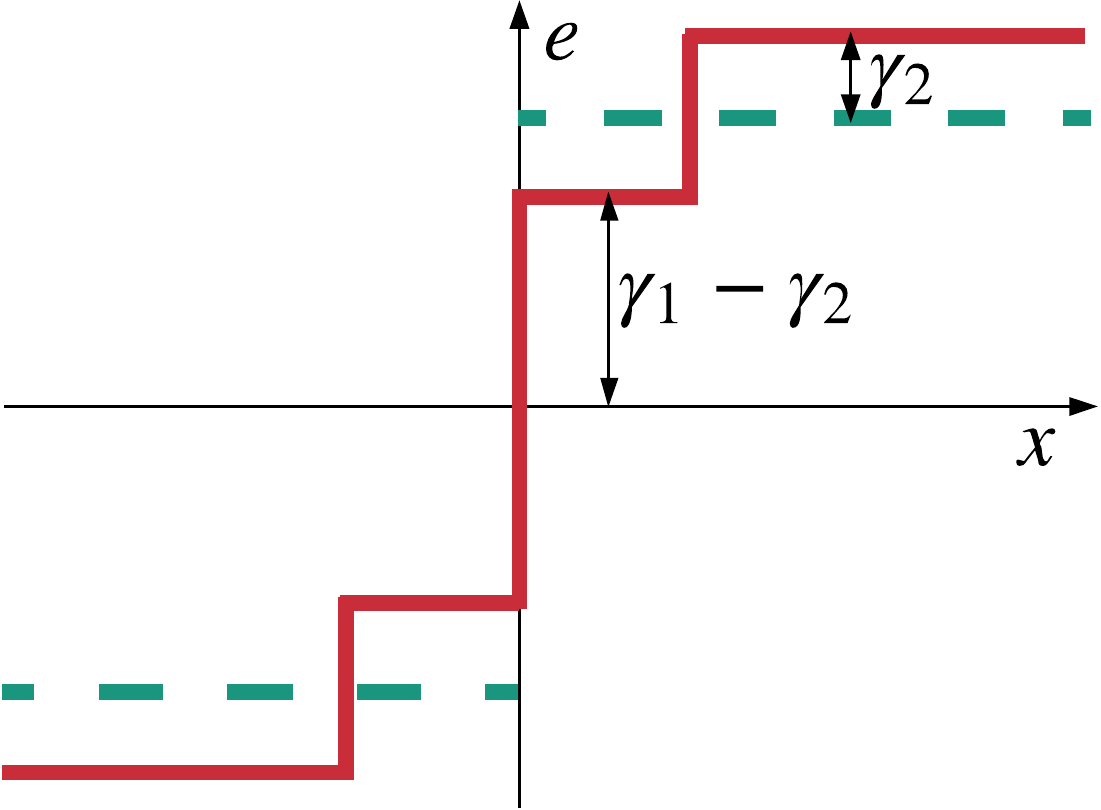}
                \caption{}
        \end{subfigure}
        \caption{Illustration of binarized activation function. (a) Conventional 1-level binarization. (b) Residual binarization with two levels. Note that the $\gamma$ parameters are universal across all features (activations) of a particular layer.}
        \label{fig:act_bin}
        \squeezeup
        \squeezeup
\end{figure}

\noindent{\bf Residual binary activation function}: Similar to previous works which use the \textit{Sign} function as the activation function, in this paper we use the residual binarization. The difference between our approach and the single-bit approach is illustrated in Figure~\ref{fig:act_bin}. The activation function includes a set of scaling factors $\{\gamma_1,\gamma_2,\dots,\gamma_M\}$ that should be learned during the training phase.

\noindent{\bf Multi-level XnorPopcount}: In \sys{}, the dot product of an $M$-level residual-binarized feature vector $\vec{e}$ and a vector of binary weights $\vec{w}$ can be rendered using $M$ subsequent $XnorPopcount$ operations. Let $\vec{e}=\sum_{i=1}^{M}{\gamma_{e_i}\vec{s}_{e_i}}$ and $\vec{w}=\gamma_w\vec{s}_w$, where $\vec{s}_{e_i}$ denotes the $i$-th residual sign vector of the features and  $\vec{s}_w$ is the sign vector of the weight vector. The dot product between $\vec{e}$ and $\vec{w}$ is computed as:
\begin{equation}
{\renewcommand{\arraystretch}{1.5}
    \begin{tabular}{ll}
         $dot(\vec{w},\vec{e})$ &$=\ dot(\sum_{i=1}^{M}{\gamma_{e_i}\vec{s}_{e_i}},\gamma_w\vec{s}_w)$  \\
         &$=\ \sum_{i=1}^{M}{\gamma_{e_i} \gamma_w\ dot(\vec{s}_{e_i}, \vec{s}_w)}$\\
         &$=\ \sum_{i=1}^{M}{\gamma_{e_i} \gamma_w\ XnorPopcount(\vec{b}_{e_i}, \vec{b}_w)}$, 
    \end{tabular}\label{eq:multi_dot}
}
\end{equation}
where $\{\vec{b}_{e_i}$, $\vec{b}_w\}$ are the binary encodings corresponding to $\{\vec{s}_{e_i}, \vec{s}_w\}$, respectively. Note that the subsequent $XnorPopcount$ operations can be performed sequentially on the same hardware accelerator, providing a tradeoff between runtime and approximation accuracy.

\subsection{Training residual binary CNNs}
Training neural networks is generally performed in two steps. First, the output layer of the neural network is computed and a cost function is derived. This step is called forward propagation. In the second step, known as backward propagation, the gradient of the cost function with respect to the CNN parameters is computed and the parameters are updated accordingly to minimize the cost function. 

For binary neural networks, the forward propagation step is performed using the binary approximations of the parameters and the features. In the backward propagation step, however, the full-precision parameters are updated. Once the network is trained, the full-precision parameters are binarized and used for efficient inference.

In this section, we derive the gradients for training residual binarized parameters in CNNs. Let $\mathcal{L}$ denote the cost function of the neural network. Consider a full-precision feature (weight) $x$ approximated by a single binary value $e=\gamma \cdot Sign(x)$. The derivatives of the cost function with respect to $\gamma$ is computed as follows:

\begin{equation}
\centering
\resizebox{0.62\columnwidth}{!}{
$\frac{\partial \mathcal{L}}{\partial \gamma}= \frac{\partial \mathcal{L}}{\partial e} \times \frac{\partial e}{\partial \gamma}= \frac{\partial \mathcal{L}}{\partial e}\times Sign(x)$
}
\end{equation}
Similarly for $x$:
\begin{equation}
\resizebox{0.91\columnwidth}{!}{
$\frac{\partial \mathcal{L}}{\partial x}= \frac{\partial \mathcal{L}}{\partial e} \times \frac{\partial e}{\partial Sign(x)} \times \frac{\partial Sign(x)}{\partial x}= \frac{\partial \mathcal{L}}{\partial e}\times \gamma \times 1_{|x|\leq 1}$
}
\end{equation}
where the derivative term $\frac{\partial Sign(x)}{\partial x}$ is approximated as suggested in~\cite{courbariaux2016binarized}:
\begin{equation}
   1_{|x|\leq 1}=\begin{cases} 
      1 & |x|\leq 1 \\
      0 & otherwise 
   \end{cases} 
\end{equation}

In a multi-level binarization scheme, input $x$ is approximated as $e=\sum_i{\gamma_i \cdot Sign(r_{i})}$ with $r_i$ denoting the $i-th$ residual error. The gradients can be computed similar to the derivatives above:
\begin{equation}
    \begin{tabular}{ll}
         $\frac{\partial \mathcal{L}}{\partial \gamma_i}$&$=\ \frac{\partial \mathcal{L}}{\partial e}\times Sign(r_{i})$  \\
         $\frac{\partial \mathcal{L}}{\partial x}$&$=\ \frac{\partial \mathcal{L}}{\partial e}\times \sum_i  \gamma_i \times 1_{|r_{i}|\leq 1}$
    \end{tabular}
\end{equation}
Note that the training phase of \sys{} is performed on a floating point processor (e.g., CPU or GPU) and the parameters have full-precision values in this phase. After training, the binary approximates are loaded into the binary hardware accelerator.

\subsection{Hardware Accelerator}\label{sec:hw}
\sys{} includes a high-level synthesis library for FPGA implementation of binary CNNs. The accelerator architecture is inspired by the one described in Figure~\ref{fig:base_accelerator} but provides a paradigm shift in term of overhead and scaling properties. Unlike the previous works that only accommodate single-bit binary CNNs, our accelerator offers a reconfigurable design that enjoys residual binarization with minimal hardware modification. In this section, we discuss different components of \sys{} accelerator and compare them with those of the baseline accelerator discussed in Section~\ref{sec:prelim-hw}.

\begin{figure}
    \centering
    \includegraphics[width=0.6\columnwidth]{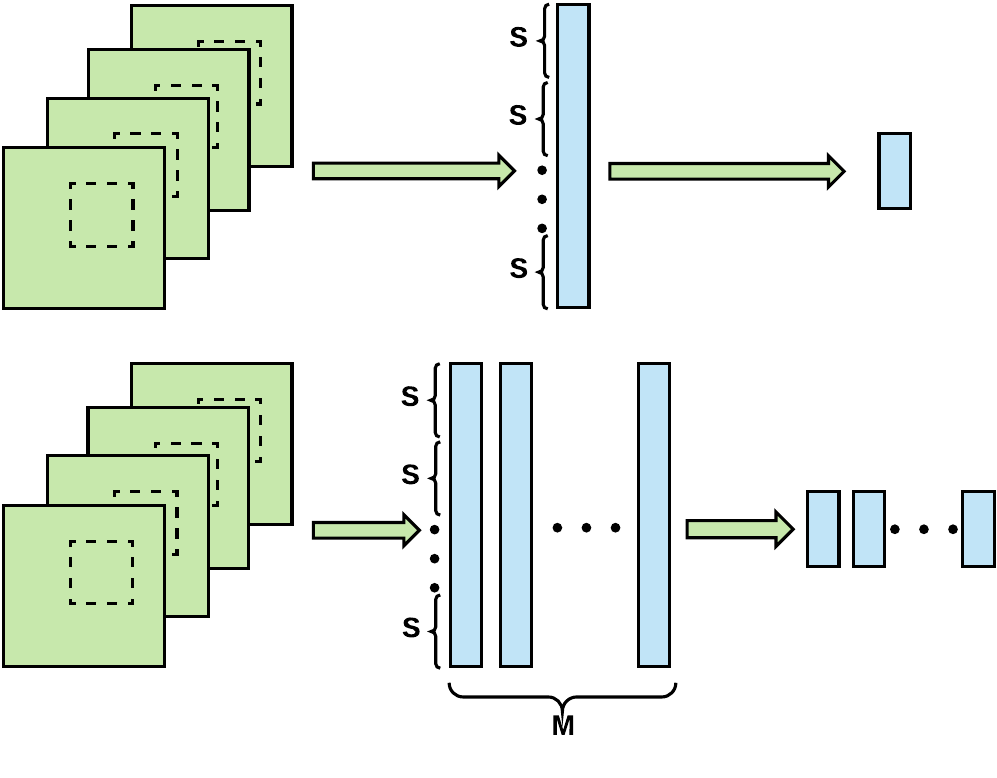}
    \caption{The baseline~\cite{umuroglu2017finn} sliding window unit (top), and our $M$-bit sliding window unit (bottom).}
    \label{fig:swu}
\end{figure}

\noindent{\bf Sliding window unit (SWU):} Figure~\ref{fig:swu} depicts a high-level overview of the SWU. It slides the (dashed) windows through the feature maps, converts them into a binary vector, and splits the binary vector into $S$-bit words ($S$ is the SIMD-width), which are sequentially sent to the MVTU. In the case of $M$-level residual binarization, the SWU still sends $S$-bit words but this time it transfers $M$ such words sequentially. This approach enables a scalable design with a fixed SIMD-width; therefore, the reconfigurability of $M$ incurs negligible hardware overhead while the runtime grows \emph{linearly} with $M$.

\begin{figure}
    \centering
    \includegraphics[width=0.98\columnwidth]{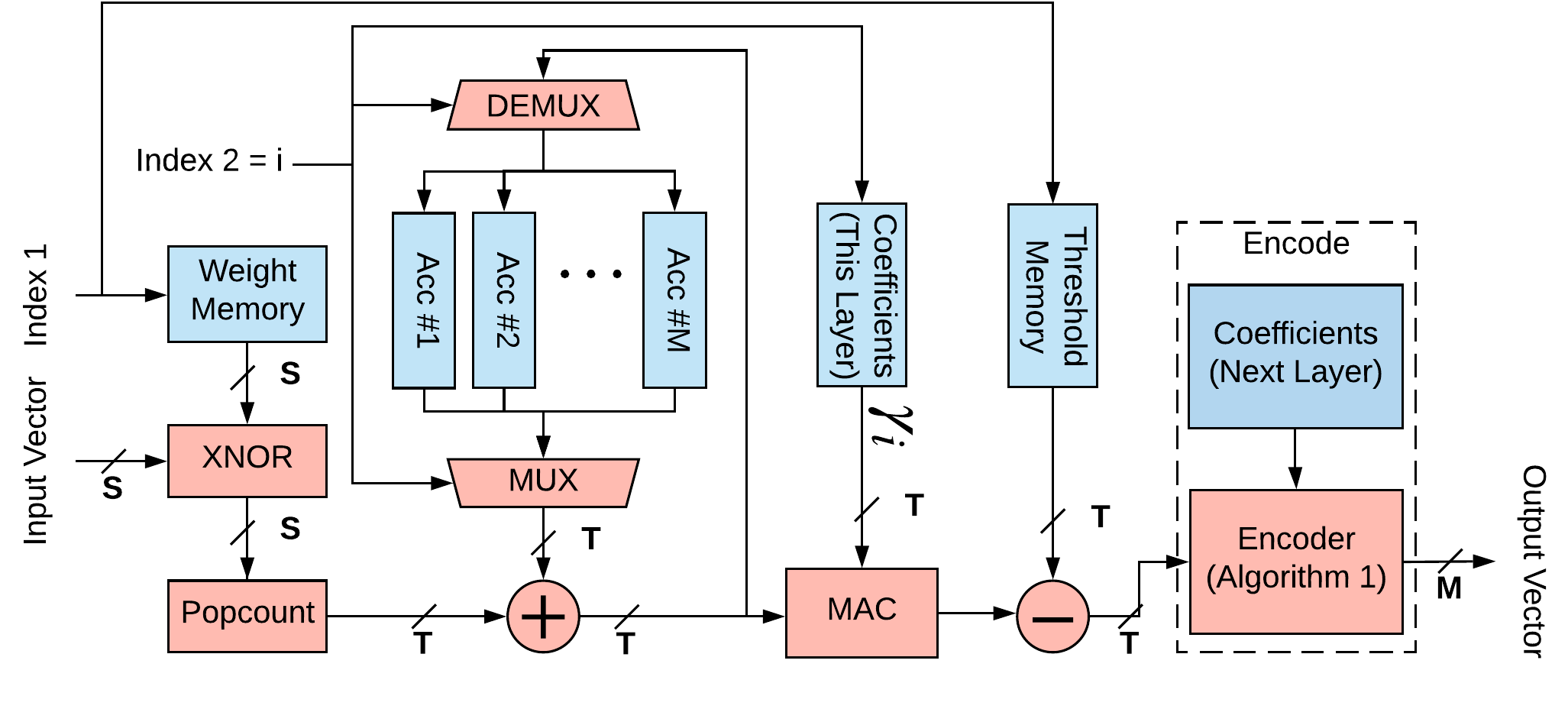}
    \caption{Architecture of processing element in \sys{}. ``S'' is the SIMD-width, ``T'' is the fixed-point bitwidth in computations, and ``M'' is the number of residual levels.}
    \label{fig:our-pe}
    \squeezeup
    \squeezeup
\end{figure}

\noindent{\bf Matrix vector threshold unit (MVTU):}
This unit is responsible for computing the neuron outputs in convolution and fully-connected layers. Similar to the baseline accelerator in Section~\ref{sec:prelim-hw}, our MVTU offers two levels of parallelism. The internal architecture of PE is shown in Figure~\ref{fig:our-pe}. Compared to the baseline unit of Figure~\ref{fig:base_accelerator}-(c), our PE maintains multiple accumulators to store the popcount values corresponding to each of the $M$ residual binarization levels. Once all $M$ popcounts are accumulated, they are multiplied by the corresponding scaling factors ($\gamma_i$) and summed together via the \textit{MAC} unit. Batch normalization is implemented using the threshold memory. Finally, the encoder module computes the $M$-bit residual representation fed to the next layer based on Algorithm~\ref{alg:resenc}. 

\noindent{\bf Max-pooling:}
The original 1-bit binarization allows us to implement max-pooling layers using simple OR operations. For $M$-level binarization, however, max-pooling layers should be implemented using comparators since performing Boolean OR over the binary encodings is no longer equivalent to performing max-pooling over the features. Nevertheless, the pooling operation can be performed over the encoded values directly. Assume full-precision values $e_x$ and $e_y$, with $M$-bit binary encodings $b_x$ and $b_y$, respectively. Considering ordered positive $\gamma_i$ values (i.e. $\gamma_1 > \gamma_2 > \cdots >\gamma_l > 0$), one can conclude that if $b_x < b_y$ then $e_x < e_y$; therefore, instead of comparing the original values $(e_x , e_y)$, we compare the binary encodings $(b_x , b_y)$ which require small comparators.

\section{Experiments}\label{sec:eval}
We implement our Software API using Keras~\cite{chollet2015keras} library. Our hardware API is implemented in Vivado HLS  and the synthesis reports (resource utilization and latency) for the FPGA accelerator are obtained using Vivado Design Suite~\cite{vivado}. We compare \sys{} with the prior art in terms of accuracy, FPGA resource utilization, execution (inference) time on the FPGA accelerator, and scalability. Proof-of-concept evaluations are performed for four datasets: CIFAR-10, SVHN, MNIST, and ILSVRC-2012 (Imagenet). Our hardware results are compared against the FINN design~\cite{umuroglu2017finn}, which uses the training method of Binarynet~\cite{courbariaux2016binarized}. The FINN paper only evaluates the first three applications. For the last dataset (Imagenet), we implemented the corresponding FINN accelerator using their open-source library. Throughout this section, $M$ denotes the number of residual binarizations and $T=24$ is the fixed-point bitwidth of features.

We implement a small network consisting only fully-connected layers for MNIST, which we call \emph{Arch-1}. The CIFAR-10 and SVHN datasets are evaluated on a medium-sized convolutional neural network named \emph{Arch-2}. Finally, the Imagenet dataset is evaluated on a relatively large network called \emph{Arch-3}. Table~\ref{tab:archs} outlines the three neural network architectures and the corresponding parallelism factors for different layers. The parallelism factors only affect hardware performance and have no effect on the CNN accuracy. For \emph{Arch-1} and \emph{Arch-2}, we set the parallelism factors exactly the same as the baselines in the FINN design. For \emph{Arch-3}, we configure the parallelism factors ourselves. Each of these architectures is implemented on a different FPGA evaluation board outlined in Table~\ref{tab:platforms}.

\begin{table}[ht!]
\centering
\caption{Network architectures for evaluation benchmarks. $C64(P,S)$ denotes a convolution with 64 output channels; the kernel size of the convolution is $3 \times 3$ and the stride is $1$ unless stated otherwise. $D512(P,S)$ means a fully-connected layer with 512 outputs. The numbers $(P,S)$ represent the two parallelism factors (PE-count, SIMD-width) for the layers.
$MP2$ stands for $2\times2$ max pooling, $BN$ represents batch normalization. Residual Binarization is shown using $RB$.}

\label{tab:archs}
\resizebox{0.85\columnwidth}{!}{
\begin{tabular}{C{0.17\columnwidth}||L{0.78\columnwidth}}
\textbf{Benchmark}&\textbf{CNN Architecture}\\ \hline 
\textbf{MNIST}&784\ (input)- D256(16,64)- BN- RB- D256(32,16)- BN- RB- D256(16,32)- BN- RB- D10(16,4)- BN- Softmax\\ \hline \hline
\textbf{CIFAR-10 \ \ \ \ \ \& \ \ \ \ \ \ \ \ \ \ SVHN}&$3\times 32\times 32$ (input)- C64(16,3)- BN- RB- C64(32,32)- BN- RB- MP2- C128(16,32)- BN- RB- C128(16,32)- BN- RB- MP2- C256(4,32)- BN- RB- C256(1,32)- BN- RB- D512(1,4)- BN- RB- D512(1,8)- BN- RB- D10(4,1)- BN- Softmax\\ \hline \hline
\textbf{Imagenet}&$3\times 224\times 224$ (input)- C96(32,33)$^*$- BN- RB- MP3- C256(64,25)$^{**}$- BN- RB- MP3- C384(64,54)- BN- RB- C384(64,36)- BN- RB- C256(64,36)- BN- RB- MP3- D4096(128,64)- BN- RB- D4096(128,64)- BN- RB- D1000(40,50)- BN- Softmax\\ 

\end{tabular}
}
\vspace{0.5em}

\scriptsize $^*$Stride is 4 for this layer, filter size is $11\times 11$.

\scriptsize $^{**}$ Filter size is $5\times 5$ for this layer.
\squeezeup
\squeezeup
\end{table}

\begin{figure*}[ht!]
    \centering
    \begin{subfigure}[b]{0.23\textwidth}
            \includegraphics[width=\columnwidth]{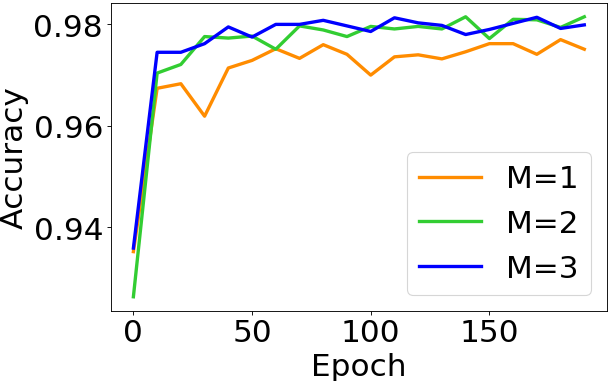}
            \caption{MNIST}
    \end{subfigure}
    ~
    \centering
    \begin{subfigure}[b]{0.23\textwidth}
            \includegraphics[width=\columnwidth]{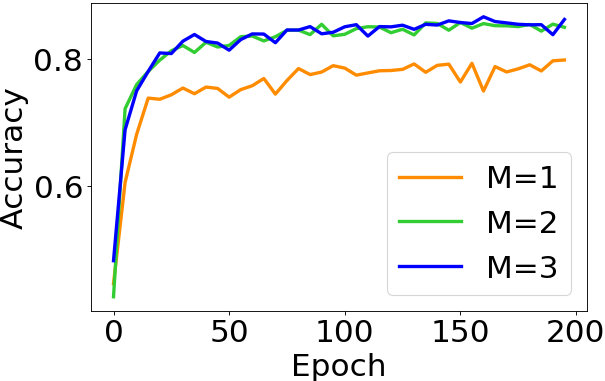}
            \caption{CIFAR-10}
    \end{subfigure}
    ~
    \centering
    \begin{subfigure}[b]{0.23\textwidth}
            \includegraphics[width=\columnwidth]{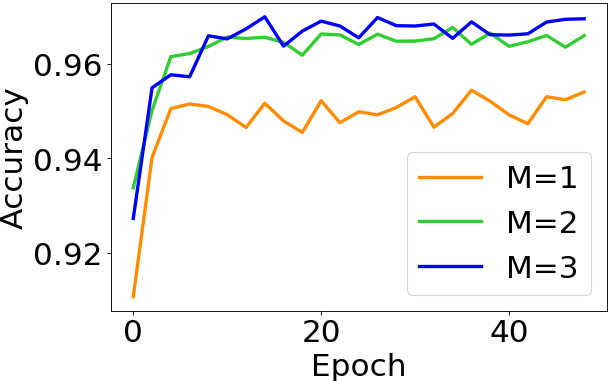}
            \caption{SVHN}
    \end{subfigure}
    ~
    \centering
    \begin{subfigure}[b]{0.23\textwidth}
            \includegraphics[width=\columnwidth]{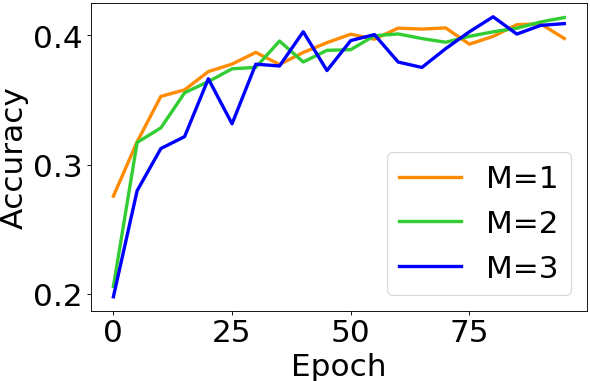}
            \caption{Imagenet}
    \end{subfigure}
    \caption{Top-1 accuracy of \sys{} for different benchmarks. $M$ denotes the number of residual binarization levels. MNIST, CIFAR-10, SVHN, and Imagenet models are trained for 200, 200, 50, and 100 epochs, respectively.}
    \label{fig:learning_curve}
    \squeezeup
\end{figure*}

\subsection{Accuracy and Throughput Tradeoff}
The accuracy of a binary CNN depends on the network architecture and the number of training epochs. The training phase is performed on a computer, not the FPGA accelerator. However, we provide the learning curve of \sys{} with different numbers of residual levels to show that residual binarization increases the convergence speed of binary CNNs.

\begin{table}
\centering
\caption{Platform details in terms of block ram (BRAM), DSP, flip-flop (FF), and look-up table (LUT) resources.}
\label{tab:platforms}
\resizebox{\columnwidth}{!}{
\begin{tabular}{|c|c|c|c|c|c|}
\hline
Application      & Platform & BRAM & DSP & FF & LUT \\ \hline
Imagenet         &   Virtex VCU108 & 3456 &  768   &      1075200 & 537600     \\ \hline
CIFAR-10 \& SVHN &   Zynq ZC702          &  280   &  220   &   106400   &  53200  \\ \hline
MNIST            &   Spartan XC7S50       &  120   &  150   &    65200  &  32600  \\ \hline
\end{tabular}
}

\end{table}

Figure~\ref{fig:learning_curve} presents the learning curves (accuracy versus epoch) for the four applications. As can be seen, increasing $M$ improves both the convergence speed and the final achievable accuracy. On the other hand, a higher $M$ requires more computation time during the execution (after the accelerator is implemented on FPGA). Table~\ref{tab:accs} compares our accuracy and throughput with the conventional binarization method proposed in~\cite{courbariaux2016binarized} and implemented (on FPGA) by the FINN design~\cite{umuroglu2017finn}. Even with 1 level of binarization, our method surpasses the FINN design in terms of accuracy, which is due to the trainable scaling factors of \sys{}. As can be seen, we can improve the accuracy of \sys{} by increasing $M$, which, in turn, reduces the throughput. We evaluate the extra hardware cost of having $M \geq 2$ in the next section.

\begin{table}
\centering
\caption{Comparison of accuracy and throughput between the FINN design and \sys{}. The baseline accuracy for the first three datasets is reported by~\cite{umuroglu2017finn}. The authors of XNOR-net~\cite{rastegari2016xnor} implement Binarynet~\cite{courbariaux2016binarized} and report the top-1 accuracy of 27.9\% for Imagenet. As $M$ grows, the accuracy of \sys{} is increased. Note that for CNN applications, even incremental accuracy gains are of importance~\cite{lecun2015deep}. The clock frequency of our designs is 200Mhz.}
\label{tab:accs}

\resizebox{0.99\columnwidth}{!}{
\begin{tabular}{ccc|c|c|c|c|}
\cline{4-7}
\multicolumn{1}{l}{}                        & \multicolumn{1}{l}{}                      &                          & MNIST            & CIFAR10        & SVHN           & Imagenet         \\ \hline
\multicolumn{2}{|c|}{\multirow{2}{*}{FINN}}                                             & Accuracy (\%)            & 95.83            & 80.1           & 94.9           & 27.9             \\
\multicolumn{2}{|c|}{}                                                                  & Throughput (samples/sec) & $6.3\times 10^5$ & $6\times 10^3$ & $6\times 10^3$ & $5.2\times 10^2$ \\ \hline
\multicolumn{1}{|c|}{\multirow{6}{*}{Ours}} & \multicolumn{1}{c|}{\multirow{2}{*}{M=1}} & Accuracy (\%)            & 97.87            & 80.59          & 95.46          & 40.89            \\
\multicolumn{1}{|c|}{}                      & \multicolumn{1}{c|}{}                     & Throughput (samples/sec) & $6.4\times 10^5$ & $6\times 10^3$ & $6\times 10^3$ & $5.3\times 10^2$ \\ \cline{2-7} 
\multicolumn{1}{|c|}{}                      & \multicolumn{1}{c|}{\multirow{2}{*}{M=2}} & Accuracy (\%)            & 98.29            & 85.94          & 96.82          & 41.37            \\
\multicolumn{1}{|c|}{}                      & \multicolumn{1}{c|}{}                     & Throughput (samples/sec) & $3.3\times 10^5$ & $3\times 10^3$ & $3\times 10^3$ & $2.6\times 10^2$ \\ \cline{2-7} 
\multicolumn{1}{|c|}{}                      & \multicolumn{1}{c|}{\multirow{2}{*}{M=3}} & Accuracy (\%)            & 98.25            & 86.98          & 97.00          & 41.43            \\
\multicolumn{1}{|c|}{}                      & \multicolumn{1}{c|}{}                     & Throughput (samples/sec) & $2.3\times 10^5$ & $2\times 10^3$ & $2\times 10^3$ & $1.7\times 10^2$ \\ \hline
\end{tabular}
\squeezeup
\squeezeup
\squeezeup
}
\end{table}

\subsection{Hardware Evaluation and Reconfigurability}

\begin{figure*}[ht!]
    \centering
    \includegraphics[width=\textwidth]{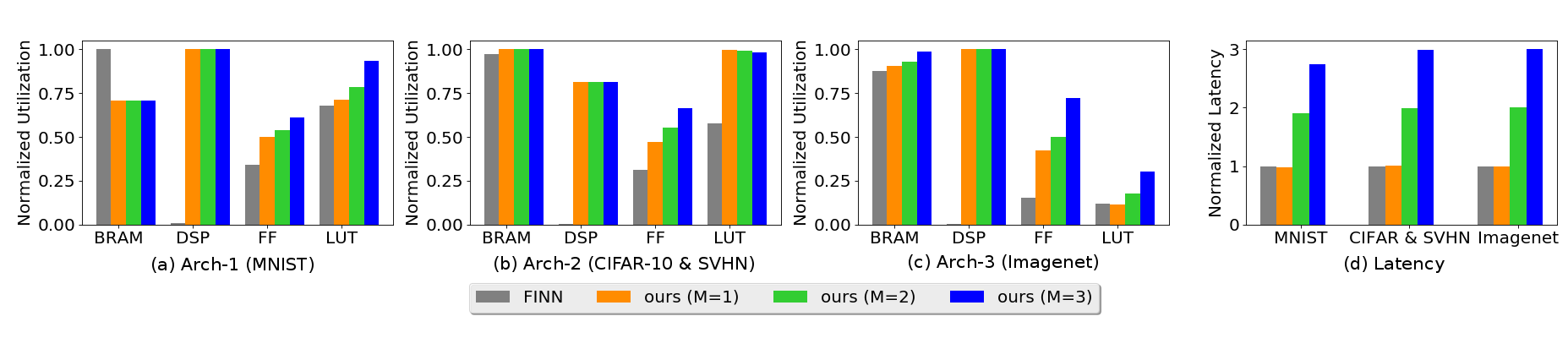}
    \caption{(a, b, c) Normalized hardware utilization and (d) latency of \sys{} compared to FINN. The utilizations are normalized by maximum resources available on each platform. The latencies are normalized by the latency of FINN. In \sys{}, the hardware overhead of reconfigurability is minimal and latency grows linearly with $M$. Note that the accuracy of FINN in all benchmarks is lower than \sys{} with $M=1$ and as we increase $M$, the accuracy is improved (see Table~\ref{tab:accs}). FINN, however, incurs excessive area and latency costs to increase the accuracy and thus is not scalable (see Figure~\ref{fig:size_effect_hw}). }
    \label{fig:hw_cost}
    \squeezeup
\end{figure*}

Figure~\ref{fig:hw_cost}-(a, b, c) compares the resource utilization of \sys{} with the baseline FINN design for each of the three network architectures. Indeed, compared to the less accurate FINN design, \sys{} has an added area cost mainly due to the design reconfigurability.
Recall the PE design in Figure~\ref{fig:our-pe}; The MAC block, the thresholding unit, and the encoder unit altogether require extra computation resources, which explains why \sys{} has extra DSP utilization compared to FINN. The BRAM, FF, and LUT utilizations of \sys{} increase with $M$ because the PEs of the design would require more accumulators and a more complex control logic. Recall that the FINN design has lower accuracy compared to \sys{} (See Table~\ref{tab:accs}). In fact, If we were to enhance the accuracy of FINN (by training wider networks), the increase in resource utilizations would exceed the limitations of the board as discussed in Section~\ref{sec:cnn_size}.

Figure~\ref{fig:hw_cost}-(d) compares the latency (runtime) of \sys{} with FINN. The latency of our accelerator with $M=1$ is the same as that of FINN; this is due to the fact that the runtime overhead of the extra modules of \sys{}, i.e., the MAC and encoder modules, is negligible compared to the $XnorPopcount$ operations. Overall, the latency of \sys{} grows linearly with $M$ since the number of $XnorPopcount$ operations (for a single dot product) is equal to $M$ (see Equation~\ref{eq:multi_dot}). As such, \sys{} offers a tradeoff between accuracy and latency.

\subsection{Effect of CNN size}\label{sec:cnn_size}
\begin{figure}
    \centering
    \includegraphics[width=0.8\columnwidth]{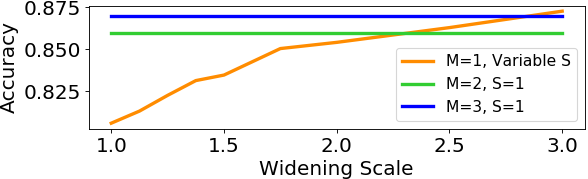}
    \caption{Effect of network widening on the accuracy of the CIFAR-10 benchmark. The widened CNNs are trained for 200 epochs and the best accuracy is accounted. Although wider single-level networks are capable of achieving the same accuracy as multi-level networks, the hardware cost and latency of wider networks are significantly larger than those of multi-level CNNs (see Figure~\ref{fig:size_effect_hw}). }
    \label{fig:size_effect_acc}
    \squeezeup
\end{figure}

As we demonstrated before, \sys{} improves the accuracy using more residual binarization levels. Another method for enhancing the accuracy is to train a \emph{wide} network as analyzed in~\cite{umuroglu2017finn}. In our notation, a reference network is widened by the scale $S$ if all hidden layers are expanded by a factor of $S$, i.e., the widened network has $S$-times more output channels in convolution layers and $S$-times more output neurons in fully-connected layers. Consider the CIFAR-10 architecture \emph{Arch-2} in Table~\ref{tab:archs}. In Figure~\ref{fig:size_effect_acc}, we depict the accuracy of the \emph{Arch-2} network with $M=1$ that is widened with different scales (variable $S$). The accuracies of the 2-level and 3-level binarization schemes correspond to the reference \emph{Arch-2} with no widening ($S=1$). It can be seen that the 1-level network should be widened by scales of 2.25 and 2.75 to achieve the same accuracy as the 2-level and 3-level networks, respectively.

\begin{figure}
     \centering
    \begin{subfigure}[b]{0.48\columnwidth}
            \includegraphics[width=\columnwidth]{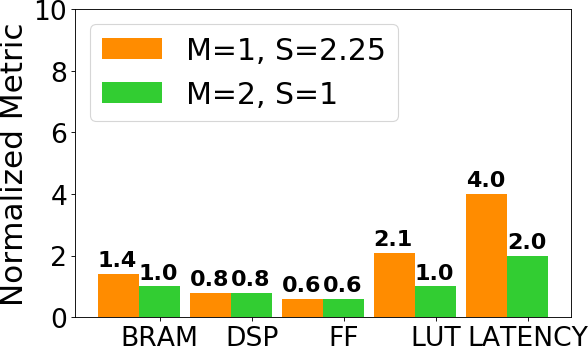}
            \caption{}
    \end{subfigure}
    ~
    \centering
    \begin{subfigure}[b]{0.48\columnwidth}
            \includegraphics[width=\columnwidth]{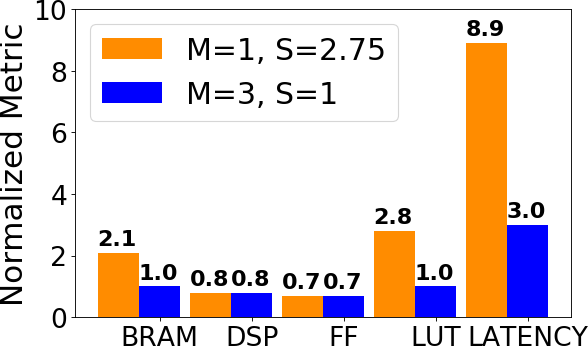}
            \caption{}
    \end{subfigure}
    \caption{Hardware cost and latency comparison of residual binarized CNN and widened CNN with the same accuracy. (a) \emph{Arch-2} widened by a scale of 2.25 compared to \emph{Arch-2} with 2-level binarization. (b) \emph{Arch-2} widened by a scale of 2.75 compared to \emph{Arch-2} with 3-level binarization. Network widening explodes the resource utlization and latency of the accelerator while \sys{} offers a scalable solution by providing a tradeoff between accuracy and latency.}
    \label{fig:size_effect_hw}
    \squeezeup
    \squeezeup
\end{figure}

In Figure~\ref{fig:size_effect_hw}, we compare the reference 2-level and 3-level residual binary networks with their widened 1-level counterparts that achieve the same accuracies. In this Figure, the resource utilization (i.e., BRAM, DSP, FF, and LUT) are normalized by the maximum resources available on the FPGA platform and the latency is normalized by that of the reference CNN (i.e., $S=1$ and $M=1$). 
It is worth noting that widening a network by a factor of $S$ increases the computation and memory burden by a factor of $S^2$ while adding to the number of residual levels $M$ incurs a linear overhead. As such, widening the network explodes the resource utilization and exceeds the platform constraints. In addition, the latency of the wide CNN increases quadratically with $S$ whereas in \sys{} the latency increases linearly with $M$. Therefore, \sys{} achieves higher throughput and lower resource utilization for a certain classification accuracy.

\subsection{Effect of Online Scaling Factors}
Authors of XNOR-net~\cite{rastegari2016xnor} suggest computing the scaling factors during the execution. Here, we show that such computations add significant hardware cost to the binary CNN accelerator. We assume that the runtime and DSP utilization overhead of online scaling factor computation are negligible. This section only considers the excessive memory footprint of online scaling factors.  
Consider a convolution layer with $K\times K$ filters and input image $I_{H\times H\times F}$ to be binarized using online scaling factors. Recall the SWU module that is responsible for generating input vectors for the MVTU. The SWU is implemented using streaming buffers, meaning that it does not store the whole input image at once. Instead, it only buffers one row of sliding windows at a time. If the features are binary, then a buffer of size $KHF$ suffices in the SWU. In the case of XNOR-net, assuming that the fixed-point values are represented with $T$ bits, the binarization operation requires two additional buffers of size $KHFT$ (for fixed-point features) and $KHT$ (for scaling factors). The ratio of the memory footprint with and without online scaling factor computation is as follows:
\begin{equation}
\resizebox{0.8\columnwidth}{!}{
    $\frac{Mem_{XNOR-net}}{Mem_{ours}}=\frac{KHF+KHFT+KHT}{KHF}=1+T+\frac{T}{F}$
    },
\end{equation}
where $T$ is the fixed-point representation bitwidth and $F$ is the number of input channels. For a single SWU, if the flip-flop and/or BRAM utilization is $P\%$, the overhead would be $(T+\frac{T}{F})P$. Table~\ref{tab:xnor-overhead} presents the estimated memory overhead for two of our architectures that include convolution layers. The overall overhead is obtained by summing the overheads corresponding to all SWUs: $\sum_{i=2}^{L}{(T+\frac{T}{F_i})P_i}$, where $L$ is the total number of convolution layers. Note that the first layer is not considered in this summation because its input is not binarized and does not need scaling factor computation.

\begin{table}
\centering
\caption{Memory utilization of the baseline binary CNN accelerator and estimated overhead of online scaling factor computation suggested by XNOR-net~\cite{rastegari2016xnor}. The bitwidth of fixed-point features is $T=24$ in our designs.}
\label{tab:xnor-overhead}
\resizebox{0.8\columnwidth}{!}{
\begin{tabular}{l|l|l|}
\cline{2-3}
                                                             & Arch2                       & Arch3                       \\ \hline
\multicolumn{1}{|l|}{FF Utilization without XNOR-net(\%)}    & 47.02                       & 42.49                       \\ \hline
\multicolumn{1}{|l|}{FF Utilization with XNOR-net (\%)}      & \textcolor{red}{373.83} & 62.69                       \\ \hline
\multicolumn{1}{|l|}{BRAM Utilization without XNOR-net (\%)} & 99.64                       & 90.42                       \\ \hline
\multicolumn{1}{|l|}{BRAM Utilization with XNOR-net (\%)}    & \textcolor{red}{236.78} & \textcolor{red}{283.47} \\ \hline
\end{tabular}
}
\squeezeup
\squeezeup
\end{table}
\section{Related Work}\label{sec:related}

Training CNNs with binary weights and/or activations has been the subject of very recent works~\cite{courbariaux2015binaryconnect,rastegari2016xnor,courbariaux2016binarized,umuroglu2017finn}. The authors of Binaryconnect~\cite{courbariaux2015binaryconnect} suggest a probabilistic methodology that leverages the full-precision weights to generate binary representatives during forward pass while in the back-propagation the full-precision weights are updated. The authors of [15] introduced binarization of both weights and activation of CNNs. The authors also suggest replacing the costly dot products by $XnorPopcount$ operations. XNOR-net~\cite{rastegari2016xnor} proposes to use scale factors during training, which results in an improved accuracy. The authors of XNOR-net do not provide a hardware accelerator for their binarized CNN. Although XNOR-net achieves higher accuracy compared to the available literature, it sacrifices the simplicity of the hardware accelerator for binary CNNs due to two reasons: (i) It utilizes multiple scaling factors for each parameter set (e.g. one scaling factor for each column of a weight matrix), which would increase the memory footprint and logic utilization. (ii) The online computation of the scaling factors for the activations requires a significant number of full-precision operations.

The aforementioned works propose optimization solutions that enable the use of binarized values in CNNs which, in turn, enable the design of simple and efficient hardware accelerators. The downside of these works is that, aside from changing the architecture of the CNN~\cite{umuroglu2017finn}, they do not offer any other reconfigurability in their designs. Providing an easily reconfigurable architecture is the key to adapting the accelerator to the application requirements.

In a separate research track, the reconfigurability of CNN accelerators has been investigated. Using adaptive low bitwidth representations for compressing the parameters and/or simplifying the pertinent arithmetic operations is investigated in~\cite{zhou2016dorefa,han2016eie,wu2016quantized}. The proposed solutions, however, do not enjoy the same simplified $XnorPopcount$ operations as in binarized CNNs. Considering the aforementioned works, the reconfigurability and scalability of binary CNN accelerators require further investigation. To the best of our knowledge, \sys{} is the first scalable binary CNN solution that embeds reconfigurability and, at the same time, enjoys the benefits of binarized CNNs. 


\section{Conclusion}\label{sec:conclusion}
This paper introduces \sys{}, a novel reconfigurable binarization scheme which aims to improve the convergence rate and the final accuracy of binary CNNs. Many existing works have tried to compensate for the accuracy loss of binary CNNs but did not consider hardware implications of their proposals. As a result, they suffer from a degraded performance and resource efficiency. We argue that a practical and scalable effort towards enhancing the accuracy of binary CNNs should not add significant fixed-point operations and/or memory overhead to the computation flow of binary CNNs. In this paper, we evaluated the hardware cost of two of the state-of-the-art methods (XNOR-net and wide-networks) for enhancing the accuracy of binary CNNs and showed that their implementation overhead is considerable. Unlike prior methods, \sys{} does not sacrifice the simplicity of the hardware architecture to achieve a higher accuracy. Our work is accompanied by an API that facilitates training and design of residual binary networks. The API is open-source to foster the research in reconfigurable machine-learning on FPGA platforms. 


\bibliographystyle{ieeetr}
{
\bibliography{mybibliography}}

\end{document}